\def\year{2020}\relax
\documentclass[letterpaper]{article} 
\usepackage{aaai20}  
\usepackage{times}  
\usepackage{helvet} 
\usepackage{courier}  
\usepackage[hyphens]{url}  
\usepackage{graphicx} 
\usepackage{booktabs}
\usepackage{subfig}
\usepackage{tabularx}
\usepackage{algorithm}
\usepackage{algorithmic}
\usepackage{amsmath}
\usepackage{amssymb}
\usepackage{float}
\usepackage{color}
\usepackage{footnote}
\makesavenoteenv{tabular}
\makesavenoteenv{table}
\usepackage{nameref}

\usepackage{graphicx}  
\title{Reranking Passages with Coarse-to-Fine Neural Retriever Enhanced by List-Context Information}
\author{
Hongyin Zhu\\
hongyin\_zhu@163.com\\
}
 
\begin{document}

\maketitle

\begin{abstract}
Passage reranking is a critical task in various applications, particularly when dealing with large volumes of documents. Existing neural architectures have limitations in retrieving the most relevant passage for a given question because the semantics of the segmented passages are often incomplete, and they typically match the question to each passage individually, rarely considering contextual information from other passages that could provide comparative and reference information. This paper presents a list-context attention mechanism to augment the passage representation by incorporating the list-context information from other candidates. The proposed coarse-to-fine (C2F) neural retriever addresses the out-of-memory limitation of the passage attention mechanism by dividing the list-context modeling process into two sub-processes with a cache policy learning algorithm, enabling the efficient encoding of context information from a large number of candidate answers. This method can be generally used to encode context information from any number of candidate answers in one pass. Different from most multi-stage information retrieval architectures, this model integrates the coarse and fine rankers into the joint optimization process, allowing for feedback between the two layers to update the model simultaneously. Experiments demonstrate the effectiveness of the proposed approach. 
\end{abstract}

\section{Introduction}
Passage reranking \cite{zhu2021collaborative} is a subtask of question answering and machine reading comprehension that involves retrieving one or several passages (text options) that can best answer a given question. Each passage contains one or several sentences, as shown in Table \ref{passage}. The most common approach is to model the question-answer (QA) pair, then compute various similarity measures between obtained representations. Finally, we can choose the high-score candidates as the answer.  

\begin{table}[!h]
\centering
\caption{An example for passage reranking}
\begin{tabularx}{0.48\textwidth}{|X|} \hline
{\bf Question}: What causes heart disease? \\
{\bf Passages}: \\
A. {\bf Cardiovascular disease} (also called {\bf heart disease}) is a class of diseases that involve the heart or blood vessels ( arteries , capillaries , and veins ). \\
B. Cardiovascular disease refers to any disease that affects the cardiovascular system , principally cardiac disease, vascular diseases of the brain and kidney, and peripheral arterial disease. \\
C. The causes of {\bf cardiovascular disease} are diverse but atherosclerosis and/or hypertension are the most common. \\
D. Additionally, with aging come a number of physiological and morphological changes that alter cardiovascular function and lead to subsequently increased risk of cardiovascular disease, even in healthy asymptomatic individuals. \\
... \\ 
{\bf Answer}: C \\ \hline 
\end{tabularx}
\label{passage}
\end{table}

Recent studies have improved the quality of general text embeddings for representing passages. BGE \cite{xiao2023c} utilizes the RetroMAE approach for pre-training and employs contrastive learning for fine-tuning. Moreover, online services such as OpenAI's text embedding \cite{neelakantan2022text} and Cohere-V3 generate text embeddings through their API. While previous work has often focused on enriching text embeddings or enhancing the interaction between question-answer pairs, they have rarely considered the influence of other candidates. As a result, the relationships between candidates have not been fully explored. When humans select an answer, they consider not only whether each individual passage aligns with the question but also the presence of superior alternatives. Especially when the passages are derived from the same document or related documents, their semantics are often incomplete, and other candidates may contain valuable information that can supplement and interpret the current passage. Enriching the representation of each passage by considering the context information from other candidates can lead to more confident results. We use "list-context" across different passages to differentiate the "context" that is often discussed in the same QA pair. Context-independent representations may limit passage semantics when other passages provide useful context information for the question. For example, passage A in Table \ref{passage} explains that ``Cardiovascular disease'' refers to heart disease. Although passage C does not mention heart disease, we can still derive relevant information from passage A.

Modeling list-context is a nontrivial task. The first challenge is to emphasize the comparative and reference information. Previous studies have tackled this challenge by utilizing hierarchical GRU-RNNs to consider context information among sentences \cite{tan2018context}. However, they did not explicitly enhance sentence representation by leveraging other candidates. Another approach is multi-mention learning, which models multiple mentions in a document to answer questions \cite{swayamdipta2017multi}. While these methods have made significant contributions, they do not explicitly model the context information in the passage list. To address this limitation, we propose a list-context attention mechanism, composed of static attention and adaptive attention. This mechanism injects list-context information into the passage, allowing each candidate to consider the whole list semantics by attending to all the candidates. Additionally, adaptive attention enables each passage to adaptively interact with other candidates by considering their correlation information. 

The second challenge is the large number of candidate passages. It's difficult to analyze thousands of passages simultaneously without running into technical issues. Previous research typically broke down a long list of passages into smaller parts and then created a context-independent representation of each sentence pair. This approach often relies on a multi-stage retrieval architecture \cite{mackenzie2018query}, where the candidate documents are repeatedly narrowed down and reordered. Our paper addresses this issue by applying a two-stage retrieval approach to the neural model. Unlike previous methods, our model streamlines the multi-stage process into a two-level (coarse-to-fine) model. First, it selects a good passage set roughly, and then it fine-tunes the selection by ignoring irrelevant instances that are far from the classification hyperplane. By training two layers of model parameters jointly, our approach enables them to collaborate and interact more effectively.

We introduce a cache policy learning (CPL) algorithm to model the two-level selection process end-to-end. The coarse selection sub-process uses a scoring function to rank the sentences in the cache memory and dynamically selects the top-k scoring sentences for further processing. In addition to the coarse selection sub-process, our model also incorporates a fine ranker to further refine the representations. Our model performs passage reranking and parameter optimization simultaneously. We conduct experiments on the WIKIQA \cite{yang2015wikiqa} and MS MARCO 2.0 \cite{nguyen2016ms} datasets. The results show the effectiveness of our approach. 
The distinctive properties of this paper are as follows: 

(i) This paper introduces the idea of enhancing passage representation by considering list-context information from other candidates.

(ii) This paper proposes a list-context attention mechanism, composed of static attention and adaptive attention, to model list-context information.

(iii) This paper introduces a C2FRetriever with a cache policy learning algorithm, which can select answers from a coarse to fine level in a single pass. The experimental results demonstrate the good performance of the proposed method. 

\section{Related Work}
Previous work employs deep learning models to enhance sentence representations and compute their similarity. 
Rocktaschel et al. \cite{rocktaschel2015reasoning} propose a textual entailment model that models word relations between sentences by using word-to-word attention on an LSTM-RNN. Severyn et al. proposes CNN\textsubscript{R} to consider overlapping words to encode relational information between question and answer. Yin et al. \cite{yin2016abcnn} propose 3 attention methods on a CNN model (ABCNN) to encode mutual interactions between sentences. Miller et al. \cite{miller2016key} proposes key-value memory networks (KV-MemeNN) to select answers by using key-value structured facts in the model memory. Wang et al. \cite{wang2017bilateral} propose a Bilateral Multi-View Matching (BiMPM) model, which utilizes an attention mechanism to model the mutual interaction of sentences at different scales. Bachrach et al.\cite{bachrach2017attention} apply two attention operations to capture more word-level contextual information, but their work still focuses on enhancing sentence-pair representations without considering list-level contextual information. A work close to ours is the hierarchical GRU-RNN Tan et al. \cite{tan2018context}, which is used to model word-level and sentence-level matching and provide a kind of contextual information. However, their approach does not explicitly enhance sentence representations by using contextual information from other candidates. Our approach incorporates list-context information to augment sentence representation. Ran et al. propose a Paragraph Comparison Network (OCN) for Multiple Choice Reading Comprehension.

Guo et al. \cite{guo2016deep} propose the Deep Dependent Matching Model (DRMM), which introduces a histogram pooling technique to summarize the translation matrix. Xiong et al. \cite{xiong2017end} propose KNRM, which uses a kernel pooling layer to softly compute the frequency of word pairs at different similarity levels. MS MARCO's official baseline \cite{mitra2017learning} uses two separate DNNs to model query-document relevance using local and distributed representations, respectively. Conv-KNRM \cite{dai2018convolutional} enhances KNRM by utilizing CNN to compose n-gram embeddings from word embeddings and cross-matched n-grams of different lengths. SAN + BERT base maintains a state and iteratively refines its predictions.

Previous work mainly uses a multi-stage retrieval architecture \cite{mackenzie2018query} in web search systems. A set of candidate documents is generated using a series of increasingly expensive machine-learning techniques. \cite{DBLP:conf/sigir/ChenGBC17} present a method for optimizing cascaded ranking models. BERT \cite{devlin2018bert} is a pre-trained language model \cite{zhu2022metaaid,zhu2023metaaid,zhu2023metaaid,zhu2023metaaid2} based on the bidirectional Transformer architecture \cite{vaswani2017attention}. It has been demonstrated to be highly effective in generating rich representations for pairs of sentences. BERT offers a means to directly model the interactions between passages. Sentence-BERT (SBERT) \cite{reimers2019sentence} is an improved model based on BERT and RoBERTa that is used to efficiently calculate the similarity between sentences. It is trained through siamese and triplet network architecture, making the representation of sentences in vector space more suitable for cosine-similarity calculations. KEPLER \cite{wang2021kepler} proposes a method to jointly optimize knowledge embedding (KE) and pre-trained language model (PLM). This means that the model simultaneously learns how to extract relational facts from the knowledge graph and how to understand language patterns from text during training. 

General text embedding techniques are commonly employed for web search and question answering, and they are also used to enhance the capabilities of large language models
\cite{zhu2021collaborative,zhu2022financial,zhu2023fqp}. 
BGE \cite{xiao2023c} adopts the RetroMAE method for pre-training and leverages contrastive learning for fine-tuning. OpenAI text embedding \cite{neelakantan2022text} produces high-quality text and code vector representations through contrastive pre-training on large amounts of unsupervised data. 
Cohere-v3\footnote{https://huggingface.co/Cohere} improves the quality of text embeddings by evaluating how well the query matches the document's topic and assessing the overall quality of the content. However, when the document is segmented, it is easy to cause semantic incompleteness. The above state-of-the-art method only enhances the text embeddings of a single passage, and cannot obtain supplementary semantic information from other passages belonging to the same document. Our method provides complementary information between passages by modeling the relationship between passages in the same document. Our C2FRetriever is trained in an end-to-end manner so that the parameters of the coarse ranker and fine ranker can be jointly optimized for better collaboration between the two levels. 

\section{Approach}
Suppose we have a question $Q$ with $l$ tokens $\{w_1^{q},w_2^{q},...,w_l^{q}\}$, and a candidate answer set $O$ with $n$ passages $\{O_1,O_2,...,O_n\}$. $n$ is the number of passages that can vary over a wide range (1$\sim$1000). Each passage $O_i$ contains one or several sentences (passage) which consists of $o_i$ tokens $\{w_1^{o},w_2^{o},...,w_{o_i}\}$. The label $y_i \in \{0,1\}$ with 1 denotes a positive answer and 0 otherwise. The goal of the model is to score each passage based on how well it answers the question, and then rank the passages based on the score. 

Our main effort lies in designing a deep learning architecture that enhances representations by considering the contextual information of other candidates. Its main building block has two layers, the coarse ranker and the fine ranker. In the following, we first describe the two layers and then describe the training algorithm. 

\begin{figure}[!h]
\centering
\includegraphics[width=2.2in]{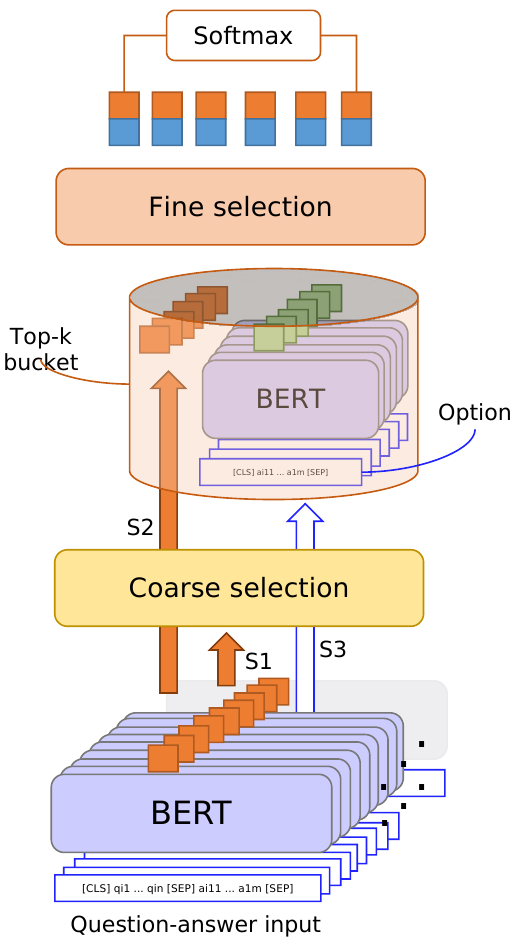}
\caption{Schematic diagram of the model architecture}
\label{overview.fig}
\end{figure}

\subsection{Coarse Ranker} 
The architecture of the coarse ranker is shown in the lower part of Figure \ref{overview.fig}. This layer aims to filter some answers that are irrelevant to the given question to generate an intermediate set for performing fine selection. We use a BERT model \cite{devlin2018bert} to generate representations of QA pairs and a single-layer neural network to compute matching scores.

Passage $O_i$ is concatenated with question $Q$ to form a complete sequence, denoted as $\langle\mbox{[CLS]};Q;\mbox{[SEP]};O_i;\mbox{[SEP]}\rangle$. 
\begin{align}
{ O}_i^{(c)} = \mathcal{F}_{bert}(\langle\mbox{[CLS]};Q;\mbox{[SEP]};O_i;\mbox{[SEP]}\rangle) 
\end{align}
where ${ O}_i^{(c)} \in \mathbb{R}^{d}$ is the QA representation. $\mathcal{F}_{bert}(\cdot)$ denotes the network defined in \cite{devlin2018bert}. The representations are then projected to match scores using a single-layer neural network.
\begin{align}
p_i^{(c)} = \sigma{(W^{c}{ O}_i^{(c)}+b^{c})} 
\end{align}
where $W^{c} \in \mathbb{R}^{d\times 1}$ and $b^{c}$ is a scalar. $\sigma(\cdot)$ is the sigmoid activation function. This layer takes all question-answer pairs as input $[(Q,O_1),(Q,O_2),...,(Q,O_n)]$, and then the model assigns a score to each candidate.

Modeling all candidates directly will result in out-of-memory issues due to the large amount of list-context information that needs to be processed simultaneously. The model parameter size is very large, and the gradient state retained during the optimization process will double the memory usage. To solve this problem, we propose using different paths and caching mechanisms. Each path represents a different function that processes the data in a way that reduces memory usage. The coarse ranker creates a cache, used to store the top-k passages it has encountered previously. According to $p_i$, passages are dynamically ranked through path S1, which contains the function listed below.
\begin{align}
{\bf h}_t=\mathcal{K}(p_i,{\bf h}_{t-1})
\end{align}
where ${\bf h}_0$ is the initial state of the empty cache. ${\bf h}_t$ is the cache state after $t$ step update and contains the selected passages. $\mathcal{K}(\cdot)$ denotes the ranking function. 

Prior early-stage models typically process all the candidates and then retain the top-k candidates which are also written to disk. Different from them, we dynamically maintain the cache which only retains the top-k scoring QA pairs where k is a hyper-parameter. The residual path S2 is used to connect the QA representation to the fine ranker, and it contains the function of the equation \eqref{path2}.
\begin{align}
\label{path2}
{\bf O}_{{\bf h}_t}^{(f)}=\mathcal{C}({\bf O}_{:t}^{(c)},{\bf h}_{t})
\end{align}
where ${\bf O}_{:t}^{(c)}$ represents the $[O_1^{(c)},O_2^{(c)},...,O_t^{(c)}]$. ${\bf O}_{{\bf h}_t}^{(f)}$ is the matrix for top-k QA pairs. $^{(f)}$ denotes the fine ranker. $\mathcal{C}(\cdot)$ is the selection function.

The residual path S3 indicates that it is used to transfer the top-k paragraphs selected in the coarse ranker to the fine ranker layer and generate a vector representation of the passage. Unlike the vector representation in the coarse ranker, it only contains the content of the passage, not the question. These passages are derived from the top k candidate paragraphs to generate list context information. The sorting result is the result of the borrowed coarse ranker without repeated calculations. This path can be described by Equation \eqref{passages}. Finally, the selected passages and representations in the cache are sent to the fine ranker.
\begin{align}
\label{passages}
{\bf P}_{{\bf h}_t}^{(f)}=\mathcal{C}({\bf P}_{:t}^{(c)},{\bf h}_t)
\end{align}
where ${\bf P}_{{\bf h}_t}^{(f)}$ is the matrix for top-k passages. ${P}_i^{(c)}$ is the representation for $i$-th passage, which can be calculated by equation \eqref{opt}.
\begin{align}
\label{opt}
{ P}_i^{(c)} = \mathcal{F}_{bert}(\langle\mbox{[CLS]};O_i;\mbox{[SEP]}\rangle) 
\end{align}

\subsection{Fine Ranker}
\label{fine.sec}
\begin{figure}[!h]
\centering
\includegraphics[width=3in]{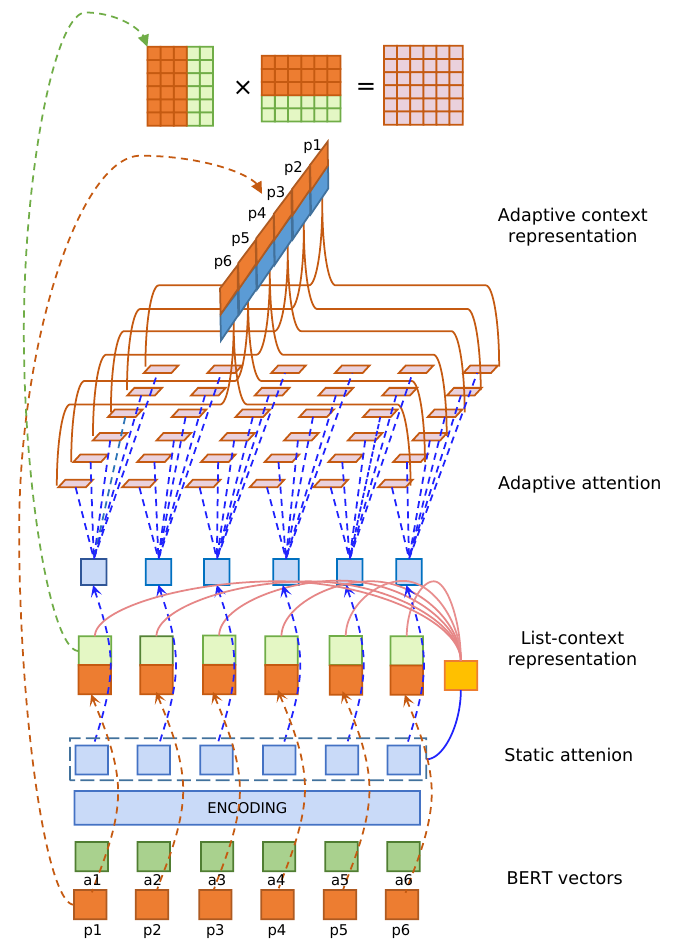}
\caption{Schematic diagram of the fine ranker}
\label{fine.fig1}
\end{figure}

The inputs to the fine ranker are vectors of top-k QA pairs ${\bf O}_{{\bf h}_t}^{(f)}$ and top-k passages ${\bf P}_{{\bf h}_t}^{(f)}$. The architecture of the fine ranker is depicted in Figure \ref{fine.fig1}. It incorporates a list-context attention mechanism that combines both static and adaptive attention.

\subsubsection{Static attention} To capture and compose the context semantics from the answer list, our model uses an attention mechanism \cite{parikh2016decomposable,yang2016hierarchical} which achieves the best performance in different alternatives to obtain the list-context representation. This method extracts informative passages and aggregates their representations to form the list-context representation $V_l$. This method first measures the weight of each passage in the list context. 
\begin{align}
\label{observer}
u_{i}&=C_l^T{P}_i^{(c)}
\end{align}
where $C_l \in \mathbb{R}^d$ is the context vector that can be jointly trained. Then, this model computes the normalized weight of each passage through a softmax function and aggregates these representations. 
\begin{align}
\alpha_{i}&=\frac{\exp(u_{i} - \max({\bf u}))}{\sum_j\exp(u_{j}-\max({\bf u}))}
\end{align}
where $\max({\bf u})$ gets the max value of $[u_0,u_1,...,u_k]$ where $k$ is the candidate number. This operation encodes the list context into a vector which summarizes the main semantics of this list. This allows the model to softly consider all candidates in the entire list.
\begin{align}
V_l&=\sum_i\alpha_{i}\cdot { P}_i^{(c)}
\end{align}
where $V_l$ is the list-context representation. 

\subsubsection{Adaptive attention} 

While the static attention supplements the list-context information for each passage, $V_l$ is the same for all candidates and does not consider the question. Ideally, we would like to overcome the above two problems, by considering list-context information and adaptively incorporating the context information for each passage based on the question. Our model resolves this problem by using adaptive attention which injects the correlation information of passages into passage representation directly, as shown in Figure \ref{fine.fig1}. This method first computes the correlation weight between these passages by considering the semantic similarity of the QA pair and the list-context information. 

\begin{align}
\label{observer}
w_{ij}&=[{{O}_i^{(c)}}^T,V_l^T] \begin{bmatrix} {O}_i^{(c)} \\ V_l \end{bmatrix}
\end{align}
Then, this method obtains normalized correlation weights through a softmax function. 
\begin{align}
\beta_{ij}&=\frac{\exp(w_{ij})}{\sum_j\exp(w_{ij})}
\end{align}
The adaptive context representation is obtained by calculating the weighted sum of the passage representations. 
\begin{align}
Z_i&=\sum_j \beta_{ij}\cdot {O}_j^{(c)}
\end{align}
where $Z_i$ is the adaptive context of $i$-th passage. By using this attention scheme, each passage has a set of adaptive correlation weights with other candidates. This correlation information can aggregate the passage representations flexibly. This interaction operation encodes the correlation between candidates while also considering the list-context information and questions. 

Then, the ranking score of the options in the cache can be calculated as below.
\begin{align}
p^{(f)}&=softmax(W^{(f)}[{\bf P}_{{\bf h}_t}^{(f)};{\bf Z}])
\end{align}
where $W^{(f)}\in \mathbb{R}^{1 \times \tau}$ and $b^{(f)}$ are linear composition matrix and bias. $\tau$ is the vector dimension. $[;]$ denotes the concatenation operation. $p^{(f)}$ is the score vector.

\subsection{Training Algorithm}

Prior multi-stage retrieval methods typically cascade different machine learning techniques. Different from the cascaded ranking architecture, we introduce the CPL algorithm to train a two-level network for the two-stage retrieval problem. This algorithm performs ranking operations during the training process and can optimize the two-stage retrieval processes jointly.

As shown in Algorithm \ref{bpl.alg}, in line 1, this model first initializes the cache memory. Then, during model training, this model adds the ground truth answer to the memory, in line 2. In lines 3-12, this model maintains the cache memory by coarsely selecting the top-k candidates. In lines 4-5, this model calculates the matching score based on the BERT representation. If the current candidate is a positive answer, its index $j$, matching score $p^{(c)}$ and representation $p_{Bert}$ will be added to the cache, in line 7. If the current candidate does not match this question, this model will maintain the top-k candidates based on their scores. In line 13, this model merges the cache memory. In line 14, this model incorporates the context information of other sentences in the fine ranker to augment the passage representation. Finally, this model calculates the loss values of coarse and fine rankers. 
Then, we can update the model parameters by back-propagation using an optimization algorithm.
\begin{algorithm}[h]
\caption{The CPL algorithm}
\label{bpl.alg}
\hspace*{\algorithmicindent} \textbf{Input}: QA pairs $pair$, Label $l$, Cache size $n$ \\
\hspace*{\algorithmicindent} \textbf{Output}: The loss values of sub-layers 

\begin{algorithmic}[1]
\STATE Initialize lists $p_i$,$p_v$,$p_{bert}$,$n_i$,$n_v$,$n_{bert}$
\STATE Get the index $p_{in}$ of positive passages
\FOR{ $ j \leftarrow 0,pair.length-1 $}
  \STATE$O_j^{(c)}=\mathcal{F}_{bert}(pair[j])$
  \STATE$p_j^{(c)}=\sigma(Linear(O_j^{(c)}))$
  \IF {$j$ in $p_{in}$}
    \STATE Update ($p_i$,$p_v$,$p_{bert}$) using ($j$,$p^{(c)}$,$O_j^{(c)}$) and maintain the score ($p^{(c)}$) order
  \ELSE
    \STATE Update ($n_i$,$n_v$,$n_{bert}$) using ($j$,$p^{(c)}$,$O_j^{(c)}$) and maintain the score ($p^{(c)}$) order
    \STATE Maintain the size of the above three ordered lists smaller than ($n$-$p_i.length$)
  \ENDIF
\ENDFOR
\STATE Merge two groups of ordered list ($p_i$,$p_v$,$p_{bert}$) and ($n_i$,$n_v$,$n_{bert}$) into cache memory ($c_i$,$c_v$,$c_{bert}$)
\STATE Generate the representations $p^{(f)}$ of fine ranker from passage list
\STATE Calculate the loss value of two selection processes $loss_c$($p^{(c)}$,$l$[$c_i$]), $loss_f$($p^{(f)}$,$l$[$c_i$])
\end{algorithmic}
\end{algorithm}

After getting the top-k sentences, this model will carry out the fine selection process as described in the \nameref{fine.sec} subsection. This model is trained using the log loss of two-level selection as shown below. 

\begin{align}
\label{obj.eq} 
\mathcal{L}_{}^{(c)}&=-\sum_{j} [y_{j}\cdot \log p_{j}^{(c)}+(1-y_{j})\cdot \log(1-p_{j}^{(c)})] \\
\mathcal{L}_{}^{(f)}&=-\sum_{j} [y_{j}\cdot \log p_{j}^{(f)}+(1-y_{j})\cdot \log(1-p_{j}^{(f)})] 
\end{align}
where $y_{j}$ is the label of the $j$-th passage. $p_{c_j}$ and $p_{f_j}$ are the predicted score of $j$-th passage in the coarse and fine rankers respectively. Then, these two layers are jointly trained to find a balance between passage selection and joint parameter optimization, as shown below.
\begin{align} 
\nonumber
\min\limits_{\theta}\mathcal{L}&=\sum_{i}^{|\mathbb{D}|}(\mathcal{L}_{i}^{(c)}+\lambda\mathcal{L}_{i}^{(f)})
\end{align}
where $\lambda$\footnote{We set $\lambda$=1.0.} is a hyper-parameter to weigh the influence of the fine ranker.

The asymptotic complexity is described as follows. Assuming that all hidden dimensions are $\rho$, the complexity of matrix ($\rho \times \rho$)-vector ($\rho \times 1$) multiplication is $O(\rho^2)$. BERT takes $O(C_{bert})$. For the coarse ranker, calculating the BERT representation of $n$ QA pairs and passages takes $O(nC_{bert})$. To maintain the cache we need a top-k sort operation which takes $O(nk)$ at the worst case. 

For the fine ranker, the list context information requires $O(\rho^2)$. The adaptive context information requires $O(k\rho^2)$. Therefore, the total complexity is $O(nC_{bert}+k\rho^2)$. For BERT, it mainly includes matrix-vector multiplication, so the optimized calculation requires $O(ml\rho^2)$, where $m$ is the number of matrix-vector multiplications, and $l$ is the sequence length. The computational complexity of this model is still close to that of BERT.

\section{Experiments}
\subsection{Datasets}
\subsubsection{WIKIQA} 
\begin{figure}[!h]
\centering
\subfloat[Training set]{
\includegraphics[width=1.8in]{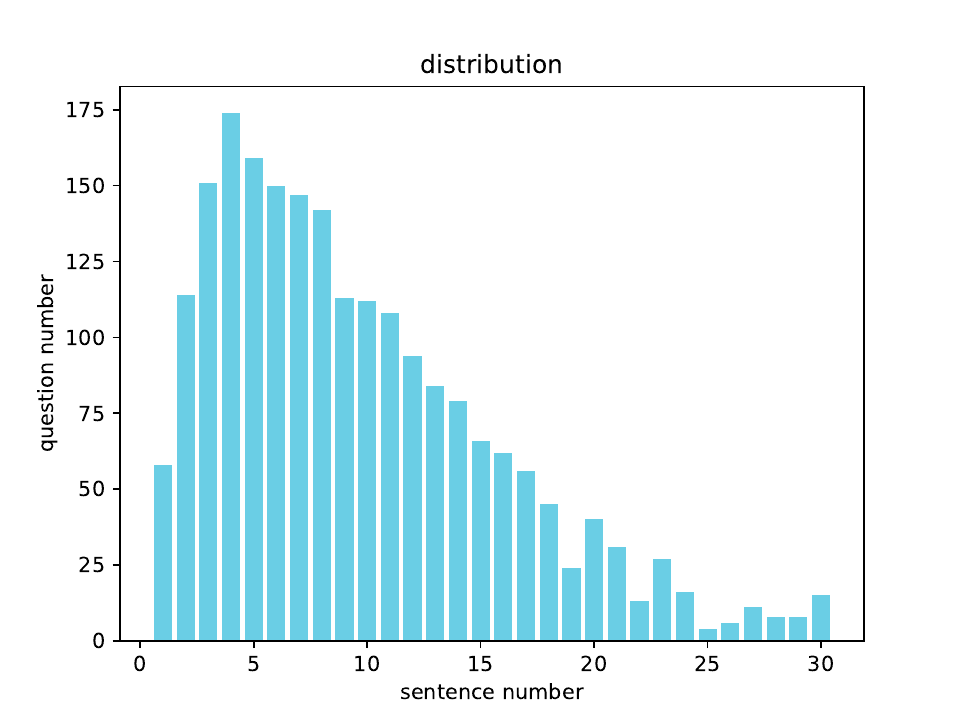}
\label{wikitrain}
}

\subfloat[Test set]{
\includegraphics[width=1.8in]{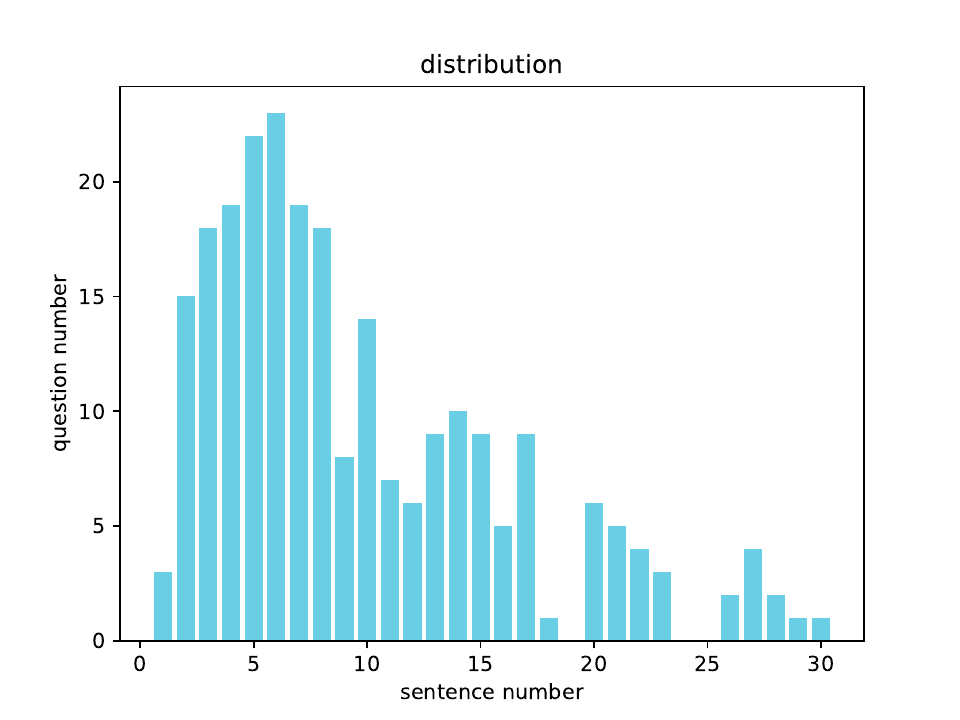}
\label{wikitest}
}
\caption{Passage distribution of the WIKIQA dataset}
\label{wikiata}
\end{figure}

This is a standard open-domain QA dataset. The questions are sampled from the Bing query logs, and the candidate sentences are extracted from paragraphs of the associated Wiki pages. This dataset includes 3,047 questions and 29,258 sentences \cite{yang2015wikiqa}, where 1,473 candidate passages are labeled as answer sentences. Each passage contains only one sentence. We use the standard data splits in experiments. Figure \ref{wikiata} visualizes the data distribution of this dataset. The x- and y- axes denote the number of candidate sentences and the number of questions respectively. The candidate number of each question ranges from 1 to 30  and the average candidate number is 9.6. The average length of question and answer are 6.5 and 25.1 words respectively.

\subsubsection{MS MARCO 2.0\footnote{https://microsoft.github.io/MSMARCO-Passage-Ranking-Submissions/leaderboard/}} 
\begin{figure}[!h]
\centering
\subfloat[Training set]{
\includegraphics[width=1.8in]{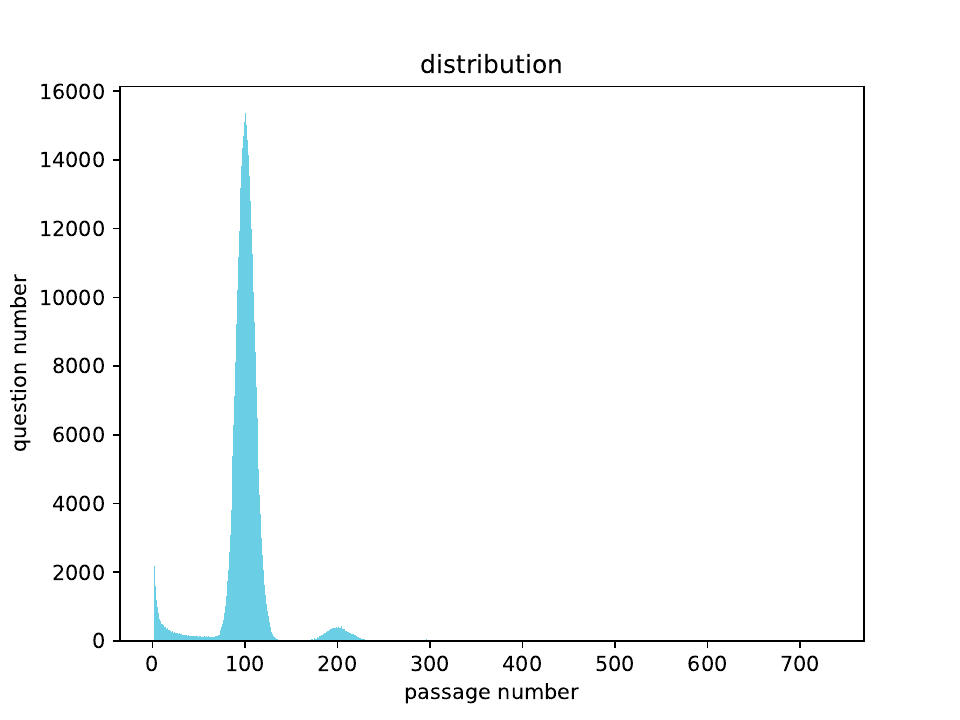}
\label{wikitrain}
}

\subfloat[Test set]{
\includegraphics[width=1.8in]{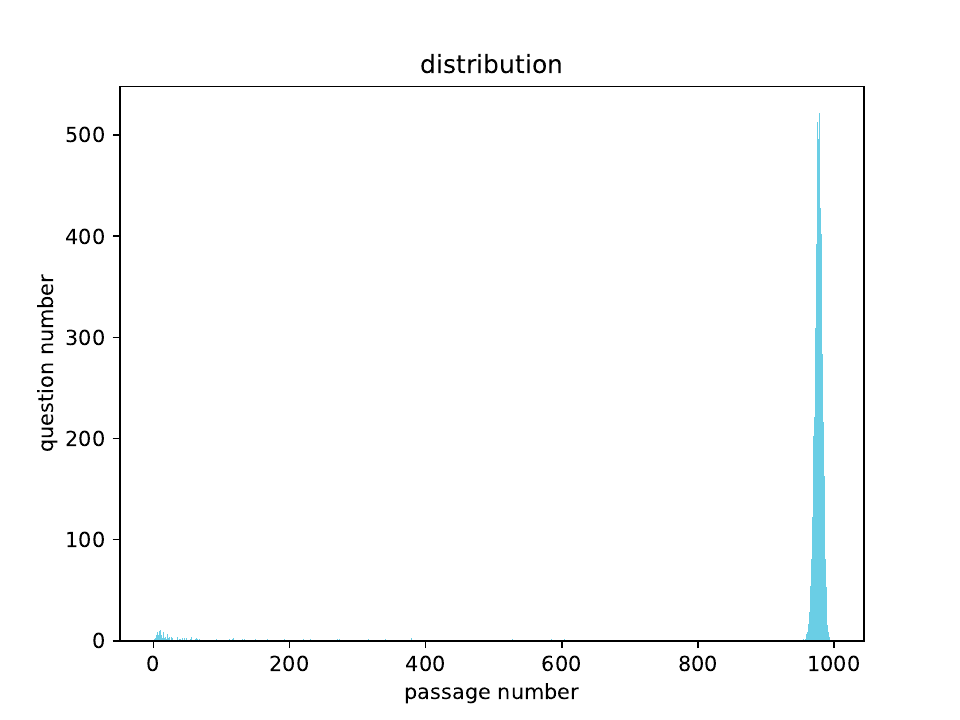}
\label{wikitest}
}
\caption{Passage distribution of the MS MARCO 2.0 dataset}
\label{marco}
\end{figure}

This is a large-scale machine reading comprehension dataset \cite{nguyen2016ms} sampled from Bing's search query logs. We choose the dataset for the passage re-ranking task. Given a set of 1000 passages that have been retrieved using the BM25 algorithm, re-rank these passages based on their relevance to the query. This dataset was the primary focus of the 2020 and 2019 TREC Deep Learning Track. It has also been utilized as a teaching resource for the ACM SIGIR/SIGKDD AFIRM Summer School, which offers courses on Machine Learning for Data Mining and Search. Figure \ref{marco} visualizes the data distribution of this dataset. The x- and y- axes denote the number of candidate passages and the number of questions respectively. The training set contains 398,792 questions. The number of passages in each question ranges from 2 to 732. The question length ranges from 1 to 38 words. The passage length ranges from 1 to 362 words. Each question average has 100.7 passage candidates. On average, each question has one relevant passage. The development set and test set contain 6,980 and 6,837 questions respectively. In the test set, the question length ranges from 2 to 30 words. The passage length ranges from 1 to 287 words. 

\subsection{Hyper-parameters}
This paper employs the bert-base-uncased model to generate representations for sentence pairs. We utilize the output of the ``[CLS]'' token from the final layer of BERT as a representation of the QA pair. The projection layer employs a single-layer neural network with the hyperbolic tangent activation function to generate 200-dimensional vector representations. We use 40 tokens as the maximum length for questions and 200 tokens for answers. The cache size is set to 15. 

The AdamW optimization algorithm \cite{loshchilov2017decoupled} is used to update the model parameters. We fine-tune the bert-base-uncased model on the passage reranking datasets. The models run on an Intel(R) Xeon(R) Platinum 8163 CPU @ 2.50GHz (Mem: 330G) \& 8 Tesla V100s and an Intel(R) Xeon(R) CPU E5-2680 v4 @ 2.40GHz (Mem: 256) \& 8 RTX 2080Tis.

\subsection{Evaluation}
This paper uses Mean Average Precision (MAP) and Mean Reciprocal Rank (MRR) to evaluate the performance of the model. For the MS MARCO 2.0 dataset, we use the official script to evaluate our results. This script calculates the MRR@10 which considers only the top 10 passages. 

\subsection{Results on the WIKIQA dataset}

\begin{table*}[!htbp]
\centering
\caption{Results on the WIKIQA dataset}
\label{wikitable}
\begin{tabular}{|l|cc|}\toprule
Method          & MAP    & MRR    \\ \midrule
WordCnt \cite{yang2015wikiqa}         & 0.4891 & 0.4924 \\
WgtWordCnt \cite{yang2015wikiqa}      & 0.5099 & 0.5132 \\
CNN-CNt \cite{yang2015wikiqa}        & 0.5170 & 0.5236 \\
CNN\textsubscript{R} & 0.6951 & 0.7107 \\
ABCNN-3 \cite{yin2016abcnn}        & 0.6921 & 0.7108 \\
KV-MemNN \cite{miller2016key}       & 0.7069 & 0.7265 \\
BiMPM \cite{wang2017bilateral}          & 0.7180 & 0.7310 \\
IARNN-Occam \cite{wang2016inner}    & 0.7341 & 0.7418 \\
CNN-MULT \cite{wang2016compare}       & 0.7433 & 0.7545 \\
CNN-CTK \cite{tymoshenko2016convolutional}        & 0.7417 & 0.7588 \\ 
wGRU-sGRU-G\textsubscript{l2}-Cnt \cite{tan2018context} & 0.7638 & 0.7852 \\ \midrule
{\bf BERT base} &  {\bf 0.7831}    &{\bf 0.7923}  \\
{\bf BERT base + MaxPooling} &  {\bf 0.8119}    &{\bf 0.8215}  \\
{\bf C2FRetriever} & {\bf 0.8448} & {\bf 0.8605} \\ \bottomrule
\end{tabular}
\end{table*}

Table \ref{wikitable} presents the experimental result. Most baseline deep learning models typically design effective feature extraction schemes to better derive features from QA pairs for calculating question-answer similarity. The BERT model has a similar goal of mapping a QA pair to a valid representation. The proposed models are the following: (i) BERT base is the simple BERT model fine-tuned on the WIKIQA dataset. The inputs to the model are all QA pairs. (ii) BERT base + MaxPooling denotes the BERT model add the max pooling. We organize the QA pairs according to the document and consider all the sentences. Note that this setting cannot tackle a situation that has unlimited candidates. The max pooling extracts the document information. Because the passage number is less than 30, so we can get the max pooling. (iii) C2Retriever is the proposed model with list and adaptive context information.

We observe that our model outperforms the BERT base and improves 1.93\% MAP and 0.71\% MRR respectively to wGRU-sGRU-G\textsubscript{l2}-Cnt \cite{tan2018context}. Considering the context information by max pooling improves 2.88\% MAP and 2.92\% MRR respectively. This means that considering the document-level context information is helpful for each passage. 
We observe that our C2FRetriever improves 6.17\% MAP and 6.82\% MRR on the BERT base model. This is because our network considers context information from other answers. The sentences of WIKIQA come from consecutive sentences in the wiki page paragraphs, so other sentences can also provide rich contextual information. Our network softly incorporates the document context information into the sentence representation.

\subsection{Results on the MS MARCO 2.0 Dataset}
\begin{table}[!htbp]
\centering
\caption{Results on the MS MARCO 2.0 dataset}
\resizebox{0.49\textwidth}{!}{
\begin{tabular}{|l|c|} \toprule
                & {MRR@10 } \\ 
Method                    & Eval          \\ \midrule
BM25      & 0.167          \\
LeToR      & 0.195          \\
Official Baseline \cite{mitra2017learning}         & 0.2517          \\
Conv-KNRM \cite{xiong2017end}              & 0.271          \\
IRNet                 & 0.281       \\
BERT base \cite{nogueira2019passage}              & 0.347\footnote{This score is generated on the development set.}          \\
BERT large \cite{nogueira2019passage}       & 0.359          \\
SAN + BERT base & 0.359         \\
Enriched BERT base + AOA index & 0.368 \\
Enriched BERT base + AOA index + CAS + Full  &  {\bf 0.393}\footnote{This score is generated with full ranking, while other models are reranking model.} \\ \midrule
{\bf C2FRetriever (200 tokens)}   &   0.347 \\ 
{\bf C2FRetriever (400 tokens)}   &   0.364 \\ \bottomrule
\end{tabular}
}
\label{raf.ranking}
\end{table}

Table \ref{raf.ranking} lists the results on the MS MARCO 2.0 dataset. Compared with the WIKIQA dataset, each passage may contain two or more sentences so the passage length varies over a wide range. The passages of any question are from different documents retrieved by a search engine, so the continuity of passages is also reduced. This experiment can better test the versatility of our method. \cite{nogueira2019passage} fine-tune the BERT base and large models and simply use the matching score of QA pairs for ranking. Note that they train their models on multiple TPUs with appropriate batch size and sequence length, which can help to better adapt the representation to the target domain. This device significantly improves model results. 

Our approach potentially enables the usage of BERT based ranking model with lower equipment requirements. However, the drawback is that we compare the model performance by truncating the passage length. We further train our model by considering longer sequences (400 tokens) with multiple GPUs. Our model achieves further improvement by 1.7\%. We achieve 0.377 on the development set, which improves 3\% from the TPU BERT base. The full-ranking method (Yan (2019)) achieves the highest score, but the drawback is that the training process is a multi-stage pipeline. In contrast, our model only uses joint training and gets the final answer in one pass. This method significantly reduces the problem's complexity.

Compared with prior works, our approach incorporates the context information of other candidates to enhance the passage representation. Each query may have hundreds of retrieved passages from a large corpus. Our network effectively integrates the coarse- and fine-selection processes by simultaneously performing model optimization and passage selection in one pass.

\subsection{Ablation Study}

\begin{table}[!h]
\caption{Model setting ablations on the WIKIQA dataset}
\label{ablation}
\centering
\begin{tabular}{|l|cc|}
\toprule
Model  & MAP & MRR \\ \midrule
C2FRetriever  & 0.8448 & 0.8605 \\ 
\ \ \ --List & 0.8236 & 0.8348  \\
\ \ \ --Adaptive & 0.8113 & 0.8215  \\
\ \ \ --Adaptive, List & 0.8009 & 0.8121 \\ 
\ \ \ --Joint training & 0.8178 & 0.8292 \\ 
\ \ \ --Adaptive, List, + MaxPooling & 0.8119 & 0.8215 \\ 
\ \ \ --Adaptive, List, + LSTM & 0.8085 & 0.8200 \\ 
\ \ \ --Adaptive, List, Two-level & 0.7831 & 0.7923 \\
\bottomrule
\end{tabular}
\end{table}

Table \ref{ablation} shows the ablation study of the effects of different model settings. A first observation is that the list context and adaptive context information are essential for a good result. Removing the List context information slightly degrades performance. This indicates that the document-level context is necessary. When we train the pipeline model, the result drops (2.7\%). This means that joint training is important for the interaction of two-level ranking.

When we replace context information with MaxPooling, the result also drops, but it is better than not considering the context. This means considering document-level context information is helpful and using MaxPooling is a straightforward choice. When we replace MaxPooling with an LSTM encoder, the result slightly drops. This is because the passages may not continuous context model and we have the long-term dependency problem because we consider all pair-wise interactions. When we remove the fine ranker, the result drops. This suggests that the two-level selection scheme is necessary. 

\begin{figure}[!h]
\centering
\includegraphics[width=3in]{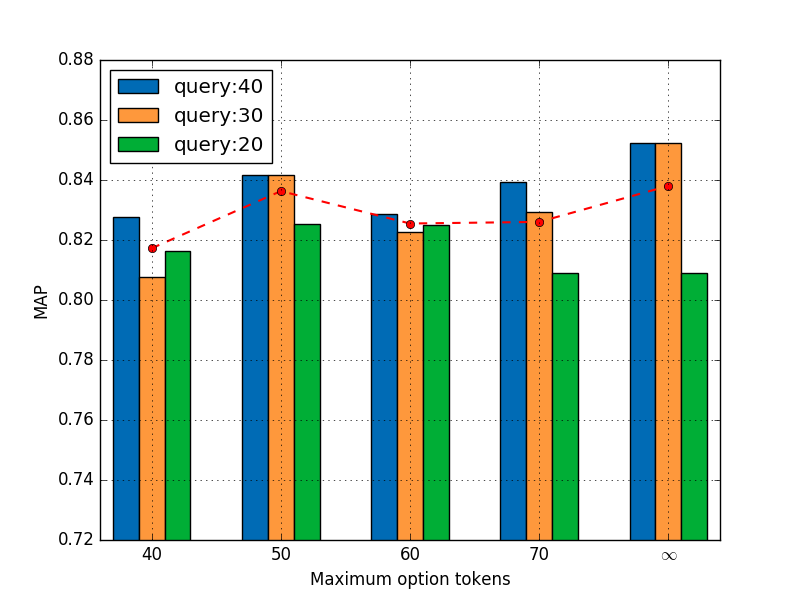}
\caption{The influence of query/passage length and passage number in fine ranker}
\label{barchat}
\end{figure}

We further analyze the impact of query/passage length and number of passages, as shown in Figure \ref{barchat}. We limit the maximum length of queries to $n$ tokens. We find that, for the same length of passage, longer query lengths generally lead to better results. However, for shorter query lengths, specifically when $n < 20$, longer passage lengths do not always result in improved outcomes. Since the meaning of the query is not well encoded, longer passages may contain more misleading information. Therefore, it is important to interpret and expand queries to improve results. We can conclude that an appropriate combination of query and paragraph lengths and model selection is crucial for the method.
%

\begin{figure}[!h]
\centering
\includegraphics[width=3in]{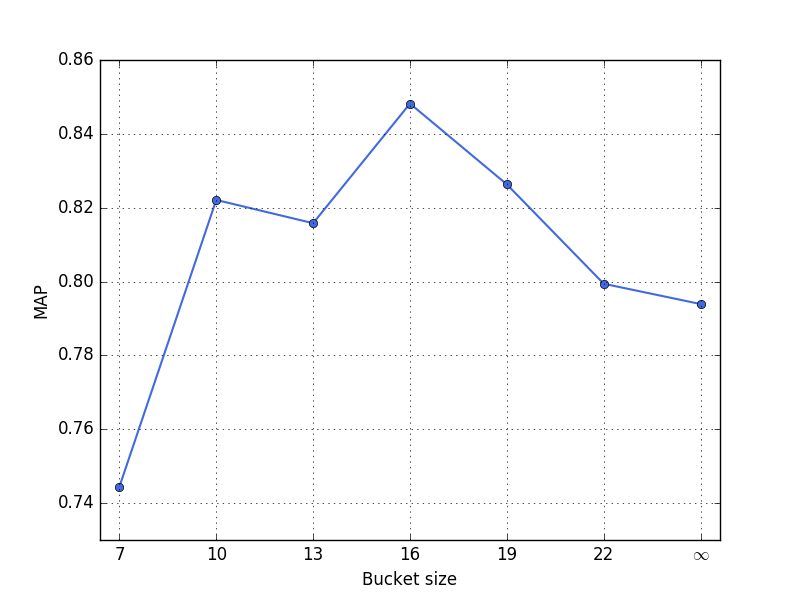}
\caption{The influence of cache size}
\label{chart}
\end{figure}
To evaluate the influence of cache size, we did extensive experiments based on different cache sizes, as shown in Figure \ref{chart}. We observe that this model achieves higher performance with the cache size=16. As the cache size increases, the performance improves as it will allow the model to consider more contextual information. When we consider many passages, it can hurt performance because it introduces noise, so considering all candidates is not always a good option. The choice of cache size is important to the result.

\section{Conclusion and Future work}
In this paper, we present a novel approach to passage reranking that incorporates list-context information to enhance the representation of passages across different contexts. Unlike previous studies, we recognize the significance of list-context information from other candidate passages in addressing the challenge of incomplete passage semantics and develop a method to integrate it effectively. Our model addresses the limitation of out-of-memory issues by leveraging a cache policy learning approach to represent list context. Additionally, we address the challenge of two-stage joint retrieval by integrating coarse and fine rankers in a seamless manner. Our model is trained by optimizing all components simultaneously, leading to the generation of the final answer in a single pass, which significantly reduces the complexity of the problem.

This paper primarily focuses on addressing the challenge of incorporating list-context information from other candidates. The model has the potential to be extended to other cascaded tasks, such as information extraction \cite{zhu2022switchnet} and downstream applications, in the future.

\bibliographystyle{aaai}
\bibliography{formatting-instructions-latex-2020}

\begin{thebibliography}{}

\bibitem[\protect\citeauthoryear{Bachrach \bgroup et al\mbox.\egroup
  }{2017}]{bachrach2017attention}
Bachrach, Y.; Zukov-Gregoric, A.; Coope, S.; Tovell, E.; Maksak, B.; Rodriguez,
  J.; McMurtie, C.; and Bordbar, M.
\newblock 2017.
\newblock An attention mechanism for neural answer selection using a combined
  global and local view.
\newblock In {\em Proceedings of ICTAI 2017}.
\newblock IEEE.

\bibitem[\protect\citeauthoryear{Chen \bgroup et al\mbox.\egroup
  }{2017}]{DBLP:conf/sigir/ChenGBC17}
Chen, R.; Gallagher, L.; Blanco, R.; and Culpepper, J.~S.
\newblock 2017.
\newblock Efficient cost-aware cascade ranking in multi-stage retrieval.
\newblock In {\em Proceedings of {ACM} {SIGIR}}.

\bibitem[\protect\citeauthoryear{Dai \bgroup et al\mbox.\egroup
  }{2018}]{dai2018convolutional}
Dai, Z.; Xiong, C.; Callan, J.; and Liu, Z.
\newblock 2018.
\newblock Convolutional neural networks for soft-matching n-grams in ad-hoc
  search.
\newblock In {\em Proceedings of WSDM}.
\newblock ACM.

\bibitem[\protect\citeauthoryear{Devlin \bgroup et al\mbox.\egroup
  }{2019}]{devlin2018bert}
Devlin, J.; Chang, M.; Lee, K.; and Toutanova, K.
\newblock 2019.
\newblock {BERT:} pre-training of deep bidirectional transformers for language
  understanding.
\newblock In {\em Proceedings of {NAACL-HLT}},  4171--4186.
\newblock Association for Computational Linguistics.

\bibitem[\protect\citeauthoryear{Guo \bgroup et al\mbox.\egroup
  }{2016}]{guo2016deep}
Guo, J.; Fan, Y.; Ai, Q.; and Croft, W.~B.
\newblock 2016.
\newblock A deep relevance matching model for ad-hoc retrieval.
\newblock In {\em Proceedings of the 25th ACM International on Conference on
  Information and Knowledge Management}.
\newblock ACM.

\bibitem[\protect\citeauthoryear{Loshchilov and
  Hutter}{2019}]{loshchilov2017decoupled}
Loshchilov, I., and Hutter, F.
\newblock 2019.
\newblock Decoupled weight decay regularization.
\newblock In {\em Proceedings of {ICLR}}.

\bibitem[\protect\citeauthoryear{Mackenzie \bgroup et al\mbox.\egroup
  }{2018}]{mackenzie2018query}
Mackenzie, J.; Culpepper, J.~S.; Blanco, R.; Crane, M.; Clarke, C.~L.; and Lin,
  J.
\newblock 2018.
\newblock Query driven algorithm selection in early stage retrieval.
\newblock In {\em Proceedings of WSDM}.
\newblock ACM.

\bibitem[\protect\citeauthoryear{Miller \bgroup et al\mbox.\egroup
  }{2016}]{miller2016key}
Miller, A.~H.; Fisch, A.; Dodge, J.; Karimi, A.; Bordes, A.; and Weston, J.
\newblock 2016.
\newblock Key-value memory networks for directly reading documents.
\newblock In {\em Proceedings of EMNLP},  1400--1409.
\newblock The Association for Computational Linguistics.

\bibitem[\protect\citeauthoryear{Mitra, Diaz, and
  Craswell}{2017}]{mitra2017learning}
Mitra, B.; Diaz, F.; and Craswell, N.
\newblock 2017.
\newblock Learning to match using local and distributed representations of text
  for web search.
\newblock In {\em Proceedings of WWW}.

\bibitem[\protect\citeauthoryear{Neelakantan \bgroup et al\mbox.\egroup
  }{2022}]{neelakantan2022text}
Neelakantan, A.; Xu, T.; Puri, R.; Radford, A.; Han, J.~M.; Tworek, J.; Yuan,
  Q.; Tezak, N.; Kim, J.~W.; Hallacy, C.; et~al.
\newblock 2022.
\newblock Text and code embeddings by contrastive pre-training.
\newblock {\em arXiv preprint arXiv:2201.10005}.

\bibitem[\protect\citeauthoryear{Nguyen \bgroup et al\mbox.\egroup
  }{2016}]{nguyen2016ms}
Nguyen, T.; Rosenberg, M.; Song, X.; Gao, J.; Tiwary, S.; Majumder, R.; and
  Deng, L.
\newblock 2016.
\newblock {MS} {MARCO:} {A} human generated machine reading comprehension
  dataset.
\newblock In {\em Proceedings of the Workshop on Cognitive Computation:
  Integrating neural and symbolic approaches 2016 co-located with the 30th
  Annual Conference on Neural Information Processing Systems {(NIPS} 2016),
  Barcelona, Spain, December 9, 2016}, volume 1773 of {\em {CEUR} Workshop
  Proceedings}.
\newblock CEUR-WS.org.

\bibitem[\protect\citeauthoryear{Nogueira and Cho}{2019}]{nogueira2019passage}
Nogueira, R., and Cho, K.
\newblock 2019.
\newblock Passage re-ranking with bert.
\newblock {\em arXiv preprint arXiv:1901.04085}.

\bibitem[\protect\citeauthoryear{Parikh \bgroup et al\mbox.\egroup
  }{2016}]{parikh2016decomposable}
Parikh, A.~P.; T{\"{a}}ckstr{\"{o}}m, O.; Das, D.; and Uszkoreit, J.
\newblock 2016.
\newblock A decomposable attention model for natural language inference.
\newblock In {\em Proceedings of {EMNLP}},  2249--2255.
\newblock The Association for Computational Linguistics.

\bibitem[\protect\citeauthoryear{Reimers and
  Gurevych}{2019}]{reimers2019sentence}
Reimers, N., and Gurevych, I.
\newblock 2019.
\newblock Sentence-bert: Sentence embeddings using siamese bert-networks.
\newblock In {\em Proceedings of {EMNLP-IJCNLP}},  3980--3990.
\newblock Association for Computational Linguistics.

\bibitem[\protect\citeauthoryear{Rockt{\"{a}}schel \bgroup et al\mbox.\egroup
  }{2016}]{rocktaschel2015reasoning}
Rockt{\"{a}}schel, T.; Grefenstette, E.; Hermann, K.~M.; Kocisk{\'{y}}, T.; and
  Blunsom, P.
\newblock 2016.
\newblock Reasoning about entailment with neural attention.
\newblock In Bengio, Y., and LeCun, Y., eds., {\em Proceedings of {ICLR}}.

\bibitem[\protect\citeauthoryear{Swayamdipta, Parikh, and
  Kwiatkowski}{2018}]{swayamdipta2017multi}
Swayamdipta, S.; Parikh, A.~P.; and Kwiatkowski, T.
\newblock 2018.
\newblock Multi-mention learning for reading comprehension with neural
  cascades.
\newblock In {\em Proceedings of {ICLR}}.
\newblock OpenReview.net.

\bibitem[\protect\citeauthoryear{Tan \bgroup et al\mbox.\egroup
  }{2018}]{tan2018context}
Tan, C.; Wei, F.; Zhou, Q.; Yang, N.; Du, B.; Lv, W.; and Zhou, M.
\newblock 2018.
\newblock Context-aware answer sentence selection with hierarchical gated
  recurrent neural networks.
\newblock {\em IEEE/ACM Transactions on Audio, Speech, and Language Processing}
  26(3):540--549.

\bibitem[\protect\citeauthoryear{Tymoshenko, Bonadiman, and
  Moschitti}{2016}]{tymoshenko2016convolutional}
Tymoshenko, K.; Bonadiman, D.; and Moschitti, A.
\newblock 2016.
\newblock Convolutional neural networks vs. convolution kernels: Feature
  engineering for answer sentence reranking.
\newblock In {\em Proceedings of the 2016 Conference of the North American
  Chapter of the Association for Computational Linguistics: Human Language
  Technologies},  1268--1278.

\bibitem[\protect\citeauthoryear{Vaswani \bgroup et al\mbox.\egroup
  }{2017}]{vaswani2017attention}
Vaswani, A.; Shazeer, N.; Parmar, N.; Uszkoreit, J.; Jones, L.; Gomez, A.~N.;
  Kaiser, {\L}.; and Polosukhin, I.
\newblock 2017.
\newblock Attention is all you need.
\newblock In {\em Advances in Neural Information Processing Systems}.

\bibitem[\protect\citeauthoryear{Wang and Jiang}{2017}]{wang2016compare}
Wang, S., and Jiang, J.
\newblock 2017.
\newblock A compare-aggregate model for matching text sequences.
\newblock In {\em Proceedings of {ICLR}}.
\newblock OpenReview.net.

\bibitem[\protect\citeauthoryear{Wang \bgroup et al\mbox.\egroup
  }{2021}]{wang2021kepler}
Wang, X.; Gao, T.; Zhu, Z.; Zhang, Z.; Liu, Z.; Li, J.; and Tang, J.
\newblock 2021.
\newblock Kepler: A unified model for knowledge embedding and pre-trained
  language representation.
\newblock {\em Transactions of the Association for Computational Linguistics}
  9:176--194.

\bibitem[\protect\citeauthoryear{Wang, Hamza, and
  Florian}{2017}]{wang2017bilateral}
Wang, Z.; Hamza, W.; and Florian, R.
\newblock 2017.
\newblock Bilateral multi-perspective matching for natural language sentences.
\newblock In {\em Proceedings of {IJCAI}},  4144--4150.
\newblock ijcai.org.

\bibitem[\protect\citeauthoryear{Wang, Liu, and Zhao}{2016}]{wang2016inner}
Wang, B.; Liu, K.; and Zhao, J.
\newblock 2016.
\newblock Inner attention based recurrent neural networks for answer selection.
\newblock In {\em Proceedings of ACL 2016}.

\bibitem[\protect\citeauthoryear{Xiao \bgroup et al\mbox.\egroup
  }{2023}]{xiao2023c}
Xiao, S.; Liu, Z.; Zhang, P.; and Muennighof, N.
\newblock 2023.
\newblock C-pack: Packaged resources to advance general chinese embedding.
\newblock {\em arXiv preprint arXiv:2309.07597}.

\bibitem[\protect\citeauthoryear{Xiong \bgroup et al\mbox.\egroup
  }{2017}]{xiong2017end}
Xiong, C.; Dai, Z.; Callan, J.; Liu, Z.; and Power, R.
\newblock 2017.
\newblock End-to-end neural ad-hoc ranking with kernel pooling.
\newblock In {\em Proceedings of the 40th International ACM SIGIR conference on
  research and development in information retrieval},  55--64.
\newblock ACM.

\bibitem[\protect\citeauthoryear{Yang \bgroup et al\mbox.\egroup
  }{2016}]{yang2016hierarchical}
Yang, Z.; Yang, D.; Dyer, C.; He, X.; Smola, A.; and Hovy, E.
\newblock 2016.
\newblock Hierarchical attention networks for document classification.
\newblock In {\em Proceedings of NAACL 2016}.

\bibitem[\protect\citeauthoryear{Yang, Yih, and Meek}{2015}]{yang2015wikiqa}
Yang, Y.; Yih, W.; and Meek, C.
\newblock 2015.
\newblock Wikiqa: {A} challenge dataset for open-domain question answering.
\newblock In M{\`{a}}rquez, L.; Callison{-}Burch, C.; Su, J.; Pighin, D.; and
  Marton, Y., eds., {\em Proceedings of {EMNLP}},  2013--2018.
\newblock The Association for Computational Linguistics.

\bibitem[\protect\citeauthoryear{Yin \bgroup et al\mbox.\egroup
  }{2016}]{yin2016abcnn}
Yin, W.; Sch{\"u}tze, H.; Xiang, B.; and Zhou, B.
\newblock 2016.
\newblock Abcnn: Attention-based convolutional neural network for modeling
  sentence pairs.
\newblock {\em Transactions of the Association for Computational Linguistics}
  4:259--272.

\bibitem[\protect\citeauthoryear{Zhu \bgroup et al\mbox.\egroup
  }{2021}]{zhu2021collaborative}
Zhu, H.; Tiwari, P.; Ghoneim, A.; and Hossain, M.~S.
\newblock 2021.
\newblock A collaborative ai-enabled pretrained language model for aiot domain
  question answering.
\newblock {\em IEEE Transactions on Industrial Informatics} 18(5):3387--3396.

\bibitem[\protect\citeauthoryear{Zhu \bgroup et al\mbox.\egroup
  }{2022}]{zhu2022switchnet}
Zhu, H.; Tiwari, P.; Zhang, Y.; Gupta, D.; Alharbi, M.; Nguyen, T.~G.; and
  Dehdashti, S.
\newblock 2022.
\newblock Switchnet: A modular neural network for adaptive relation extraction.
\newblock {\em Computers and Electrical Engineering} 104:108445.

\bibitem[\protect\citeauthoryear{Zhu}{2022a}]{zhu2022financial}
Zhu, H.
\newblock 2022a.
\newblock Financial data analysis application via multi-strategy text
  processing.
\newblock {\em arXiv preprint arXiv:2204.11394}.

\bibitem[\protect\citeauthoryear{Zhu}{2022b}]{zhu2022metaaid}
Zhu, H.
\newblock 2022b.
\newblock Metaaid: A flexible framework for developing metaverse applications
  via ai technology and human editing.
\newblock {\em arXiv preprint arXiv:2204.01614}.

\bibitem[\protect\citeauthoryear{Zhu}{2023a}]{zhu2023fqp}
Zhu, H.
\newblock 2023a.
\newblock Fqp 2.0: Industry trend analysis via hierarchical financial data.
\newblock {\em arXiv preprint arXiv:2303.02707}.

\bibitem[\protect\citeauthoryear{Zhu}{2023b}]{zhu2023metaaid}
Zhu, H.
\newblock 2023b.
\newblock Metaaid 2.0: An extensible framework for developing metaverse
  applications via human-controllable pre-trained models.
\newblock {\em arXiv preprint arXiv:2302.13173}.

\bibitem[\protect\citeauthoryear{Zhu}{2023c}]{zhu2023metaaid2}
Zhu, H.
\newblock 2023c.
\newblock Metaaid 2.5: A secure framework for developing metaverse applications
  via large language models.
\newblock {\em arXiv preprint arXiv:2312.14480}.

\end{thebibliography}

\end{document}